# Generating Unobserved Alternatives

Shichong Peng
Simon Fraser University &
University of California San Diego
shichong_peng@sfu.ca

Ke Li
Simon Fraser University, Google &
Institute for Advanced Study
keli@sfu.ca

## Abstract

*We consider problems where multiple predictions can be considered correct, but only one of them is given as supervision. This setting differs from both the regression and class-conditional generative modelling settings: in the former, there is a unique observed output for each input, which is provided as supervision; in the latter, there are many observed outputs for each input, and many are provided as supervision. Applying either regression methods and conditional generative models to the present setting often results in a model that can only make a single prediction for each input. We explore several problems that have this property and develop an approach that can generate multiple high-quality predictions given the same input. As a result, it can be used to generate high-quality outputs that are different from the observed output.*

## 1. Introduction

Supervised learning is centred around prediction. In the classification or regression setting, only a single label/target is assumed to be correct, and the goal is predict the label with high confidence or generate a prediction that is as close as possible to the target. In settings such as multi-label prediction or class-conditional generative modelling, there could be *multiple* prediction targets for the same input that are all correct. For example, in class-conditional generative modelling, the input is the class label and all data points that belong to that class are correct prediction targets. Multiple prediction targets for the same input are given as supervision, and the goal is to generate *all* such prediction targets for the same input (class label).

In this paper, we consider a different problem setting with the following properties: (1) for the same input, there could be *multiple* prediction targets that are correct, but (2)

Results and code are available at https://niopeng.github.io/HyperRIM/.

only a single prediction target per input is given as supervision. The goal is still to generate all prediction targets for the same input. See Table 1 for a comparison of the problem setting we consider to other common settings. Note that we focus on the case of continuous prediction targets and leave discrete labels to future work.

When do such prediction problems arise? They often come up in inverse problems, which require generating *more* information from *less* information, information that is not present in the input. For example, consider the problem of super-resolution, which aims to generate a high-resolution image from a low-resolution image. The high-frequency details are completely missing from the low-resolution image, but they must be generated in the high-resolution image.

Inverse problems are typically *ill-posed*, that is, the input cannot uniquely determine the output and so there could be multiple valid outputs for the same input. However, only one of them is actually observed. Concretely, in the case of super-resolution, there are many ways to generate details in the high-resolution image, and the observed high-resolution image used for training represents only one of these ways. This combination of *one-to-many* prediction and *one-to-one* supervision characterizes the problem setting we consider.

The problem essentially requires us to generate alternatives that were never observed, so a natural question is why it should be possible at all. After all, if there were a valid alternative output that was never realized, how do we know whether it exists, and why should the model generate such an alternative if there is no indication that it exists? The answer lies in an observation that holds true across many natural problems: *which* of the many valid prediction targets is observed is usually arbitrary, and so while a valid alternative for the current input may not be observed, we expect an analogous version of it for *some* other input to be observed. Therefore, the hope is for the model to generalize across different inputs to produce the full range of alternative predictions for all inputs.

In this paper, we take an initial step towards addressing this problem and propose an approach for it based on Implicit Maximum Likelihood Estimation (IMLE) [26]. We demonstrate on three problems that the approach can produce different alternative predictions for the same input, even though only one prediction target is given for each input.

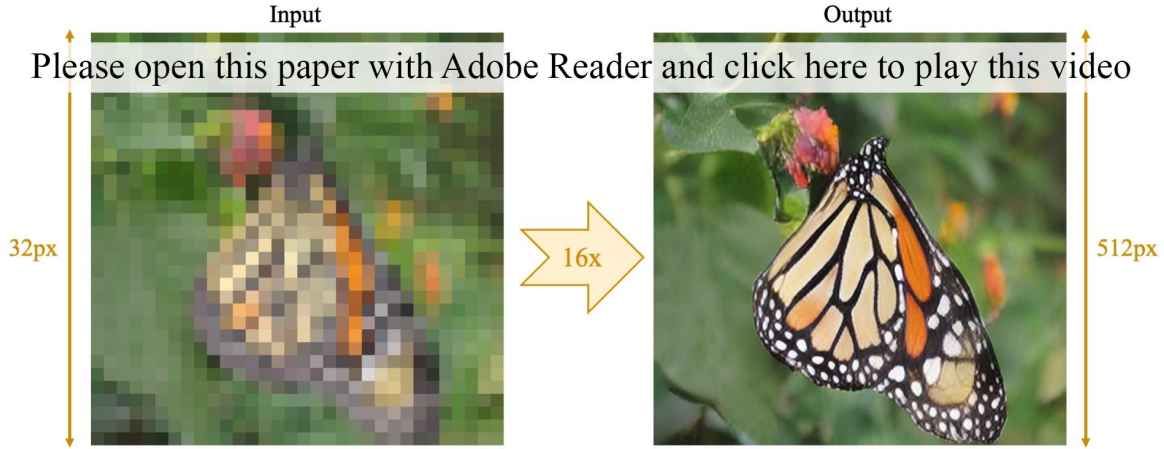

Figure 1: Example unseen input image and output from our method (HyperRIM) demonstrated on super-resolution, which upscales the input image by a factor of 16, i.e.: increasing both the width and height by a factor of 16, resulting in an increase in the number of pixels by a factor of 256. Video also available as ancillary file on arXiv.

| Problem Setting | Label Type | Prediction | Supervision |
|---|---|---|---|
| Regression | Continuous | One-to-one | One-to-one |
| Classification | Discrete | | |
| Class-conditional Generative Modelling | Continuous | One-to-many | One-to-many |
| Multi-label Prediction | Discrete | | |
| Present Setting | Continuous | One-to-many | One-to-one |

Table 1: Comparison of the problem setting we consider to other common settings.

## 2. An Illustrative Example using MNIST

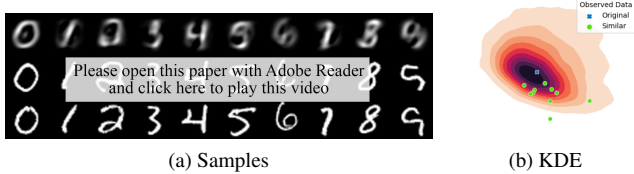

(a) Samples  (b) KDE

Figure 2: Example unseen input digits and outputs from our method. Top row is the input, middle row is the predictions and bottom row is the original images. Video also available as ancillary file on arXiv.

To illustrate the problem setting, we will start with a simple illustrative example using MNIST. We consider the problem of predicting from the first ten principal components of a data point the values of the remaining ones. More concretely, we perform principal component analysis (PCA) and project each data point onto the PCA basis. The input is the image reconstructed from the first ten coordinates and the observed output is the original image.

This prediction problem is inherently one-to-many, but only one-to-one supervision is available. Specifically, given the first ten coordinates of a real data point, there are many possible ways to fill in the values of the remaining coordinates that will result in plausible MNIST digits. However, only one of these is observed, namely the original real data point.

To illustrate what the unobserved alternatives could be, we visualize the results of our method (the details of which will be discussed later) in Figure 2a. All the predictions share the same first ten coordinates, but differ in the remaining ones. As shown, all predictions are plausible, but differ from the original images.

We can visualize the marginal distribution over the 11th and 12th coordinates of the predictions and compare to those of the real data point. As shown in Figure 2b, the real data point lies in a high density region of the prediction distribution, suggesting the method is able to predict the real data point (or at least the 11th and 12th coordinates). Note that there is only a *single* data point we can observe for the given input, because other data points in the dataset have different coordinates along the first 10 principal components and therefore differ from the given input.

As a proxy for other data points that *could* have been observed for the given input, we visualize ten data points whose first 10 principal components are the *closest* to the given input. While they technically do not match the given input (because the first 10 principal components are different from the given input), they are hopefully similar to unobserved alternatives and can therefore give us a sense of how the unobserved alternatives would be distributed. As shown, the prediction distribution has moderately high density at most of these points, indicating that they can be predicted by the method.

## 3. Method

One-to-many prediction problems can be naturally formulated in probabilistic terms. If we use $\mathbf{x}$ to denote the input, $\mathbf{y}$ to denote the prediction, our goal is to learn $p(\mathbf{y}|\mathbf{x})$. Ideally $p(\mathbf{y}|\mathbf{x})$ should assign high probability density to both observed and unobserved valid predictions, and low probability density elsewhere. So, each mode of $p(\mathbf{y}|\mathbf{x})$ corresponds to a valid prediction.

Regression models take the form of a deterministic function from $\mathbf{x}$ to $\mathbf{y}$, and so $p(\mathbf{y}|\mathbf{x})$ is always a delta. In order to produce non-deterministic predictions, the most direct way to extend regression models is to add a latent random variable as an input to the deterministic function. More precisely, a prediction is given by $\mathbf{y} \coloneqq T_\theta(\mathbf{x}, \mathbf{z})$ where $\mathbf{z} \sim \mathcal{N}(0, \mathbf{I})$. This is variously known as an implicit generative model [30], a neural sampler [32] or a decoder-based model [47].

Such a model can be trained as a conditional GAN, where $T_\theta(\cdot, \cdot)$ is interpreted as the generator. In practice, due to mode collapse, some valid predictions cannot be produced by the generator. This problem is exacerbated in the presently considered setting with one-to-one supervision: since there is only one observed output $\mathbf{y}$ for each input $\mathbf{x}$, there is only one mode to collapse to. As a result, all samples of the generator conditioned on the same input $\mathbf{x}$ are identical and the random variable $\mathbf{z}$ is effectively ignored. Hence, the generator becomes a deterministic mapping from $\mathbf{x}$ to $\mathbf{y}$, akin to a vanilla regression model.

To obtain non-deterministic predictions $\mathbf{y}$ despite the availability of only a single observation, we propose training the model using Implicit Maximum Likelihood Estimation (IMLE), which avoids mode collapse, as illustrated in Figure 4.

### 3.1. Implicit Maximum Likelihood Estimation (IMLE)

Implicit Maximum Likelihood Estimation (IMLE) [26] is a method for training implicit generative models. Compared to GANs, there are two differences: it explicitly aims to cover all modes, and optimizes a non-adversarial objective. To achieve the former, IMLE reverses the direction in which generated samples are matched to real data: rather than making each generated sample similar to some real data point, it makes sure each real data point has a similar generated sample. To achieve the latter, it removes the discriminator (which matches generated samples to real data implicitly) and instead explicitly performs matching using nearest neighbour search. The latter can be done efficiently using DCI [24, 25], which avoids the curse of dimensionality.

More precisely, if we denote the generator parameterized by $\theta$ as $T_\theta(\cdot)$, which takes in a random code $\mathbf{z}_j$ and outputs a sample, IMLE optimizes the following objective:

$$\min_\theta \mathbb{E}_{\mathbf{z}_1,\ldots,\mathbf{z}_m \sim \mathcal{N}(0,\mathbf{I})} \left[ \sum_{i=1}^n \min_{j \in \{1,\ldots,m\}} d(T_\theta(\mathbf{z}_j), \mathbf{y}_i) \right],$$

where $\mathbf{y}_i$ is a real data point, $d(\cdot, \cdot)$ is a distance metric and $m$ is a hyperparameter.

### 3.2. Conditional IMLE

IMLE can be extended to model conditional distributions by separately applying IMLE to each member of a family of distributions $\{p(\mathbf{y}|\mathbf{x}_i)\}_{i=1}^n$. If we denote the generator as $T_\theta(\cdot, \cdot)$, which takes in an input $\mathbf{x}_i$ and a random code $\mathbf{z}_{i,j}$ and outputs a sample from $p(\cdot|\mathbf{x}_i)$, the method optimizes the following objective:

$$\min_\theta \mathbb{E}_{\mathbf{z}_{1,1},\ldots,\mathbf{z}_{n,m} \sim \mathcal{N}(0,\mathbf{I})} \left[ \sum_{i=1}^n \min_{j \in \{1,\ldots,m\}} d(T_\theta(\mathbf{x}_i, \mathbf{z}_{i,j}), \mathbf{y}_i) \right],$$

where $\mathbf{y}_i$ is the observed output that corresponds to $\mathbf{x}_i$, $d(\cdot, \cdot)$ is a distance metric and $m$ is a hyperparameter. We use LPIPS perceptual distance [52] as our distance metric. Algorithm 1 shows the conditional IMLE training procedure.

### 3.3. Model Architecture

Different types of generative models require different architectures due to differences in behaviour (e.g.: mode seeking vs. covering) and training dynamics (e.g.: adversarial vs. non-adversarial) [42, 41, 36]. In this paper, we introduce a new architecture for IMLE which substantially outperforms prior IMLE architectures [26, 27]. As we will show later, this is critical to generating high quality images.

The model architecture relies on a backbone consisting of two branches. The first branch mainly consists of a sequence of residual-in-residual dense blocks (RRDB) [43], which is a sequence of three dense blocks (Fig. 6b) connected by residual connections (Fig. 6a). The second branch consists of a mapping network [19] produces a scaling factor and an offset for each of the feature channels after each RRDB in the first branch. Additionally we added weight normalization [38] to all convolution layers. While various design motifs are inspired by other works, combining them in a way that gave good performance when trained with IMLE was non-trivial and required thorough experimentation. We found the optimal hyperparameter settings to differ substantially between GAN-based and IMLE-based architectures. For example, we reduced the number of RRDB blocks by a factor of 4 and substantially expanded the number of channels compared to ESRGAN. What is new is not the design motifs themselves, but the development of an architecture for IMLE that can generate high-quality images. We expect this to be of practical interest in broader contexts, because this architecture combined with IMLE can offer benefits that cannot be obtained with other methods, such as training stability, mode coverage, fast sampling and high-quality samples.

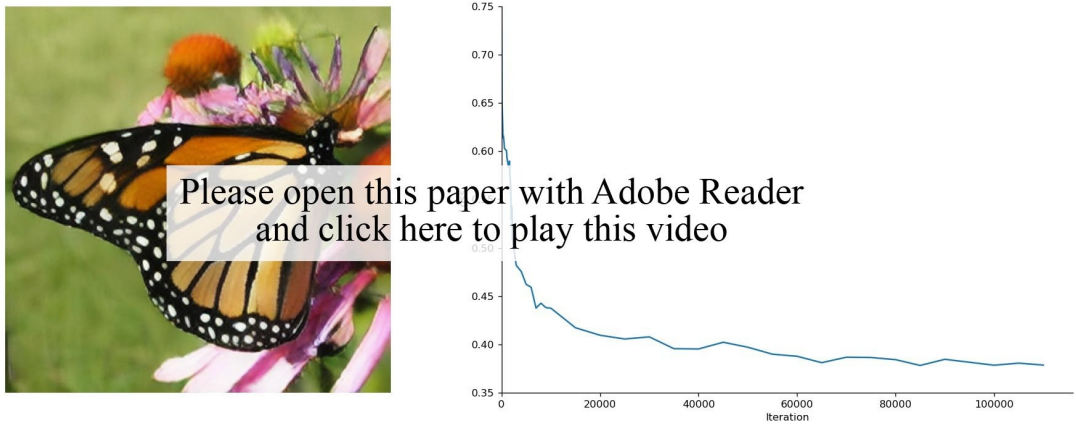

Figure 3: Click on the image to see output of model while it trains, demonstrating stable training. Video also available as ancillary file on arXiv.

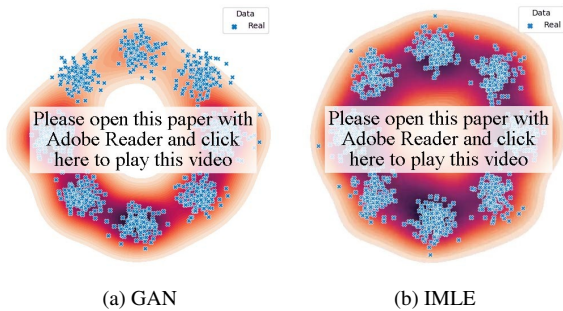

(a) GAN        (b) IMLE

Figure 4: Comparison of the extent of mode coverage attained by GAN and IMLE. Real data points are represented as blue crosses and the probability density of generated samples is visualized as a heatmap. Click on the figures above to play videos, which show the density of generated samples over the course of training. As shown, (a) fails to assign high probability density to the three clusters of real data points near the top, (b) assigns high probability density to all clusters of real data points. This shows that IMLE is able to cover all modes, whereas GAN drops some modes (near the top). Videos also available as ancillary files on arXiv.

**Algorithm 1** Conditional IMLE Training Procedure
---
**Require:** The set of inputs $\{\mathbf{x}_i\}_{i=1}^n$ and the set of corresponding observed outputs $\{\mathbf{y}_i\}_{i=1}^n$
Initialize the parameters $\theta$ of the generator $T_\theta$
**for** $p = 1$ **to** $N$ **do**
    Pick a random batch $S \subseteq \{1, \ldots, n\}$
    **for** $i \in S$ **do**
        Randomly generate i.i.d. $m$ latent codes $\mathbf{z}_1, \ldots, \mathbf{z}_m$
        $\tilde{\mathbf{y}}_{i,j} \leftarrow T_\theta(\mathbf{x}_i, \mathbf{z}_j) \; \forall j \in [m]$
        $\sigma(i) \leftarrow \arg\min_j d(\mathbf{y}_i, \tilde{\mathbf{y}}_{i,j}) \; \forall j \in [m]$
    **end for**
    **for** $q = 1$ **to** $M$ **do**
        Pick a random mini-batch $\tilde{S} \subseteq S$
        $\theta \leftarrow \theta - \eta \nabla_\theta \left( \sum_{i \in \tilde{S}} d(\mathbf{y}_i, \tilde{\mathbf{y}}_{i,\sigma(i)}) \right) / |\tilde{S}|$
    **end for**
**end for**
**return** $\theta$

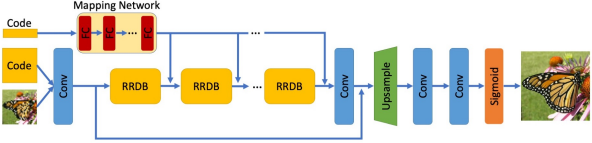

Figure 5: Details of the architecture backbone. See Figure 6a for the inner workings of RRDB blocks.

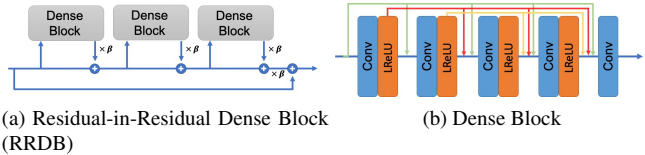

(a) Residual-in-Residual Dense Block (RRDB)

(b) Dense Block

Figure 6: (a) Inner workings of Residual-in-Residual Dense Blocks (RRDBs), which comprises of dense blocks (details in (b)). $\beta$ is the residual scaling parameter. (b) Inner workings of dense blocks.

## 4. $16\times$ Super-Resolution

Single-image super-resolution (SISR) is a classic problem in image processing. Applications span consumer and industrial use cases, and range from photo enhancement to medical imaging. Most methods consider moderately low upscaling factors (e.g. $2-4\times$). We consider an upscaling factor of $16\times$, where the width and height are both increased by 16 times, and so the number of pixels is increased by 256 times. Under this setting, the input contains much less information about the output, and so there could be a lot more valid output images for the same input image [3]. The problem therefore represents an ideal testbed for our method.

### 4.1. Progressive Upscaling

We adopt an approach of progressive upscaling, where we upscale the image by 2 times at a time. We chain together four backbone architectures which become sub-networks in a larger architecture, as shown in Figure 7. Each sub-network takes a latent code and the output of the previous sub-network, or if there is no previous sub-network, the input image.

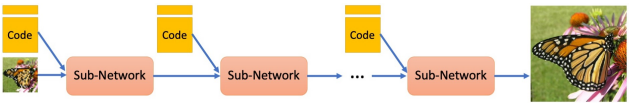

Figure 7: Our HyperRIM model consists of multiple sub-networks, each of which upscales by a factor of 2 and concatenates a random code to its input.

We add intermediate supervision to the output of each sub-network, so that the distance metric in IMLE is chosen to be the sum over LPIPS distances between the output of each sub-network and the original image downsampled to the same resolution.

Additionally, we use a hierarchical sampling procedure to generate the pool of samples IMLE operates over. Because conditional IMLE only uses the sample that is most similar to the original image for backpropagation, we can improve the sample efficiency by sampling only in the region likely to be close to the original image, which can be viewed as a way of increasing the effective number of samples. To this end, we generate a set of latent codes for the first sub-network and select the latent vector that corresponds to the sub-network output that results in a sample that is most similar to the downsampled original image. Then for each subsequent sub-network, we fix the latent codes for all previous sub-networks and generate a set of latent codes only for the current sub-network, effectively drawing samples conditioned on the selected latent codes for lower resolutions. Note that this procedure is only used at train time; at test time, the latent code for each sub-network is independently drawn.

### 4.2. Experimental Setting

In order to generate multiple versions of high-resolution images, multiple instances of the same object category must be observed. Therefore, we chose a dataset curated by semantic category, i.e.: ILSVRC-2012, and selected a subset of three categories consisting of 3900 images. To obtain the input and target output images, we downsampled them anisotropically to $512 \times 512$ and $32 \times 32$ respectively using a bicubic filter. The train and test images are disjoint.

We compare our method, HyperRIM, to the leading GAN-based and IMLE-based methods, namely ESRGAN [43] and SRIM [27] (differences with our method are contrasted in the appendix). ESRGAN is a conditional GAN trained with the relativistic GAN objective [18] and also uses two auxiliary losses on raw pixels and VGG features. Since ESRGAN was originally designed for $4\times$ upscaling, we stack two separate ESRGAN models to upscale the input image by $16\times$. To make the generator capable of producing non-deterministic predictions, we concatenate a random code to the inputs of both models.

### 4.3. Quantitative Results

We evaluate all methods according to two metrics, Fréchet Inception Distance (FID) [14] and faithfulness-weighted variance [27]. The former measures perceptual quality of the output images, while the latter measures the diversity of the different output images for the same input image weighted by their consistency with the original image. Traditional metrics like PSNR and SSIM are not suitable for our problem setting because they assume only one

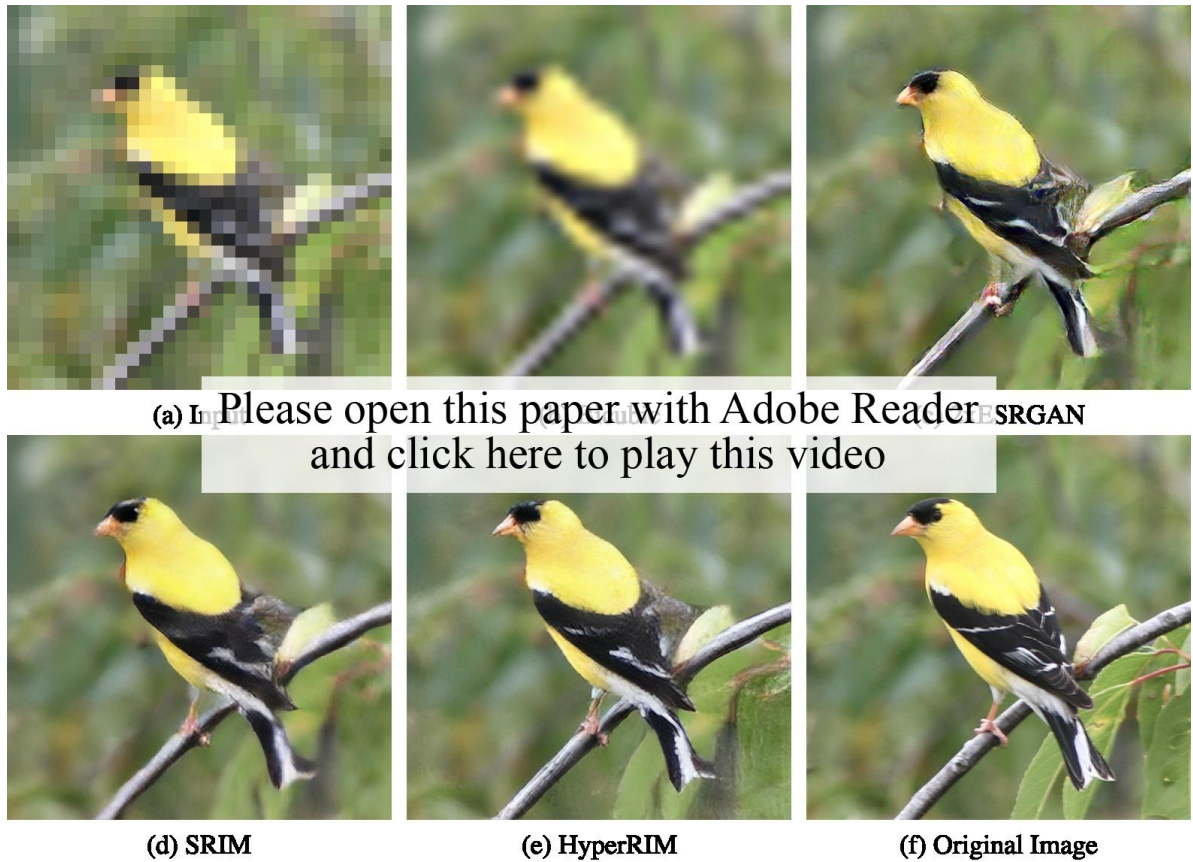

Figure 8: Visualization of different samples generated by our method (HyperRIM) and the baselines. Clicking on the figure will play video of different outputs for the same input produced by each method. As shown, (c) 2xESRGAN generates near-identical samples, (d) SRIM generates diverse samples that are less visually plausible, (e) HyperRIM generates samples that are both diverse and consistent with the input. Video of this result and results on other images also available as ancillary files on arXiv.

|  | 2xESRGAN | SRIM | HyperRIM |
|---|---|---|---|
| FID | 22.69 | 27.34 | **16.75** |

(a) Super-Resolution

|  | Pix2Pix | HyperRIM |
|---|---|---|
| FID | 110.80 | **94.84** |

(b) Image Decompression

Table 2: Comparison of Fréchet Inception Distance (FID) to the target of the samples generated by our method (HyperRIM) and the baselines. Lower values of FID are better. We compare favourably on this perceptual metric (FID).

correct target per input and do not capture perceptual similarity [23].

As shown in Table 2a, HyperRIM outperformed both baselines in terms of FID, indicating that it produces higher-quality images than both. As shown in Table 3, HyperRIM achieved higher faithfulness-weighted variance than the baselines at all bandwidth parameters. So, the outputs of our method are more diverse and consistent with the original image.

### 4.4. Qualitative Results

We show the results of our method and the baselines in Figure 8 and the appendix. As shown, HyperRIM generates better quality results than all baselines. As shown, the outputs generated by HyperRIM are more diverse than 2xES-RGAN (which produces near-identical outputs) and more realistic than SRIM.

#### 4.4.1 Training Stability

In Figure 3, we visualize the output of HyperRIM for a test input image over the course of training. As shown,

| $\sigma$ | 2xESRGAN | SRIM | HyperRIM |
|---|---|---|---|
| 0.3 | $2.87 \times 10^{-2}$ | $5.48 \times 10^{-2}$ | $\mathbf{5.72 \times 10^{-2}}$ |
| 0.2 | $1.67 \times 10^{-3}$ | $5.22 \times 10^{-3}$ | $\mathbf{5.86 \times 10^{-3}}$ |
| 0.15 | $4.83 \times 10^{-5}$ | $2.73 \times 10^{-4}$ | $\mathbf{3.44 \times 10^{-4}}$ |

Table 3: Comparison of faithfulness weighted variance of the samples generated by our method (HyperRIM) and the baselines. Higher value shows more variation in the generated samples that are faithful to the original image. $\sigma$ is the bandwidth parameter for the Gaussian kernel used to compute the faithfulness weights.

the output quality improves steadily during training, thereby demonstrating training stability.

#### 4.4.2 Precision and Recall

In Figure 9, we qualitatively evaluate the precision and recall of each method, i.e.: whether the trained model can generate (a) *only* valid outputs, and (b) *all* valid outputs. Since only images that have a corresponding latent code **z** can be generated, we can explore the space of latent codes, which should be equivalent to the space of images that can be generated. We perform the following experiment: for a test image, we optimize over the latent code to try to find an image that is as close as possible to the original high-resolution image as measured by LPIPS and visualize the images we encounter along the way. For an ideal model, traversing the space of latent codes should (a) *only* pass through valid outputs (i.e. achieves high precision), and (b) be able to reach *any* valid image, including the original image (i.e.: achieves high recall). We find that HyperRIM is able to achieve better precision and recall than the baseline.

### 5. Image Decompression

Most images are stored in a compressed format such as JPEG, and the original uncompressed images are lost. Significant artifacts may result when the images are decompressed using JPEG; to restore the original quality of images that are only stored in compressed form, it would be beneficial to learn to generate the original image from the compressed version. The input does not contain enough information to uniquely determine the output, and so it would be useful to produce multiple plausible uncompressed images and allow the user to choose one to their liking.

We choose a single backbone network as our architecture with one change: we removed the upsampling layer because the input and output resolutions are the same for decompression.

#### 5.1. Experimental Setting

To generate training data, we compressed each image from the RAISE1K [5] dataset using JPEG with a quality of 1%. We compare our method to Pix2Pix [17], given the lack of a dedicated method for image decompression [1].

#### 5.2. Results

We compare the results to the baseline in terms of FID in Table 2b. Our method, HyperRIM, achieves a lower FID than the baseline, demonstrating better perceptual quality. We visualize the outputs of our method and Pix2Pix in Figure 10. As shown, our method was able to remove most blocky artifacts, including those on the trapezoid with marble-like texture. Additionally, our method can recover different output images with different colour tones. This makes sense, because JPEG compression can cause global colour distortions.

### 6. Related Work

The proposed problem setting is related to multi-label prediction [15] and mixture regression [46]. Both aim to predict multiple targets. In the former, the labels are usually discrete and multiple labels per input are given as supervision. In the latter, while the labels are continuous, a fixed number of modes is assumed for every input.

In terms of the underlying technique, the proposed approach relies on implicit generative models, and so related are work on GANs [11, 12, 29, 33, 17] and IMLE [26, 27].

In terms of the tasks, there is a large body of work on super-resolution, most of which consider upscaling factors of $2-4\times$. See [49, 31, 45] for comprehensive surveys. Interpolation-based methods, like bilinear, bicubic and Lanczos filtering [7], compute the pixel values of the high-resolution image by interpolating between the neighbouring pixel values in the low-resolution image according to a predefined formula. Unfortunately, these methods cannot generate high-frequency details. Exemplar-based methods [9, 8, 10] sidestep this problem by copying pixels from most similar patches in a dataset. One drawback is that different patches often do not blend well. Optimization-based approaches [50, 16] try to get around this issue by performing optimization on each image to find the best way to combine the patches. This comes at the cost of increased computational burden on each new image.

Learning-based methods using deep neural nets are perhaps most popular and predict a high-resolution image from the low-resolution image using a trained model. Many methods regress to the high-resolution image directly and differ widely in the architecture [6, 20, 21, 40, 13, 53, 4, 28].

---
[1]Not to be confused with learned image compression methods, which changes the way the image is encoded. In this setting, we are given the JPEG encoded image, and so compression methods cannot be used.

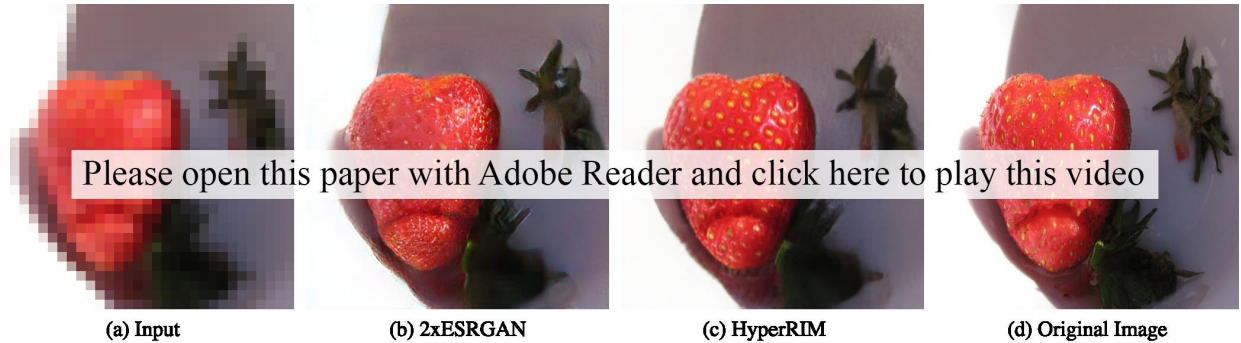

| (a) Input | (b) 2xESRGAN | (c) HyperRIM | (d) Original Image |

Figure 9: Visualization of output images from each method while traversing the space of random codes using gradient descent to reach the original image. As shown, (b) fails to reach the original image, (c) reaches the original image and only encounters images with plausible content and texture. This reveals both the (i) precision and (ii) recall of each method, i.e.: the ability of each method to generate (i) *only* plausible images and (ii) *all* plausible images, which include the original image. Since (b) cannot generate the original image, its recall is low. Since (c) can reach the original image and does so smoothly without generating an implausible image, the recall and precision of (c) are high. Video also available as ancillary file on arXiv.

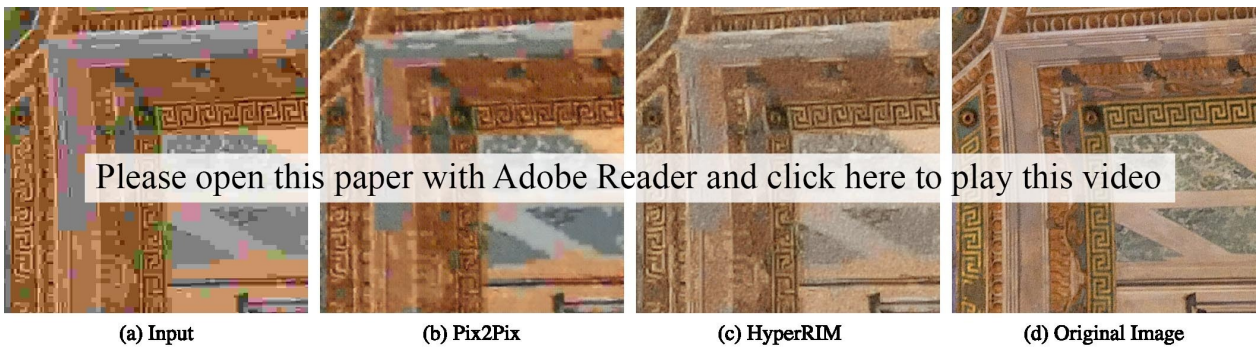

| (a) Input | (b) Pix2Pix | (c) HyperRIM | (d) Original Image |

Figure 10: Visualization of compressed input, decompressed output images from Pix2Pix and our method (HyperRIM) and the observed target image. As shown, the Pix2Pix output contains large pixel blocks whereas HyperRIM output successfully removes most artifacts. Video of this result and results on other images also available as ancillary files on arXiv.

Because these methods can only output one image for each input, they cannot handle multimodality and must compromise between different modes, which yields blurry results. Deep generative models can in principle handle multimodality by modelling a distribution over output images conditioned on the input image, and various conditional GANs have been developed for the problem [23, 37, 51, 44, 51, 35, 48, 43]. However, because conditional GANs exhibit significant mode collapse and tend to ignore the latent noise [17], they can in practice only generate a single mode. Because a mode is sharper than a compromise among modes, these methods dominate the current state-of-the-art. However, these methods can overfit to dominant patterns at the expense of less common ones and so at times generate spurious high-frequency patterns. Our method is similar in that it uses a deep generative model, but is trained using IMLE [26]. Because IMLE is non-adversarial and does not collapse modes, our method is easier to train and also handles multimodality.

Our method is also related to methods that progressively upscale the input through a number of intermediate resolutions, e.g.: [34, 22, 44]. Concurrently to this work, there has been work on extreme super-resolution which tries to upscale a fairly large image by $16\times$ [39], for which no implementation is publicly available. The challenges are however different, because the input image already contains rich structure and a fair amount of details.

There is relatively little work on image decompression to our knowledge; however, more work was done on image compression [2, 1], which changes the encoding of the compressed image itself.

## 7. Conclusion

In this paper, we considered a setting where prediction is inherently one-to-many, but where supervision is only one-to-one. This differs from traditional settings like regression or class-conditional generative modelling – in the former, both prediction and supervision are one-to-one, whereas in the latter, both are one-to-many. We explored several problems with this characteristic and demonstrated that our approach was able to generate different plausible outputs for the same input, even though only one output per input is available as supervision. Moreover, we introduced an architecture for IMLE which outperformed GAN-based methods and can offer benefits like training stability and the lack of mode collapse.

**Acknowledgements** The authors thank Kuan-Chieh Wang, Devin Guillory, Tianhao Zhang and Jason Lawrence for helpful feedback on an early draft of this paper. This research was enabled in part by support provided by WestGrid (www.westgrid.ca) and Compute Canada (www.computecanada.ca).

## A. Difference from Super-Resolution Baseline Methods

In Table 4, we contrast the conceptual differences between our method (HyperRIM) and the methods that form the basis of the baselines, namely ESRGAN [43] and SRIM [27]. We compare along the following dimensions:

- IMLE-based: Whether the model is trained using the IMLE objective
- GAN-based: Whether the model is trained using the GAN objective
- Progressive Upscaling: Whether the model generates progressively higher resolutions rather than the highest resolution at once
- Hierarchical Sampling: Whether the model employs a hierarchical sampling strategy as described in Sect. 4.1
- Intermediate Supervision: Whether the model is trained with losses on intermediate resolutions
- Weight Normalization: Whether the model uses weight normalization for all of its convolutional layers
- Has Mapping Network: Whether the model contains a separate mapping network as discussed in Sect. 3.3
- Has Pixel Loss: Whether the loss function contains a pixel-wise loss term
- Has Feature Loss: Whether the loss function contains a feature loss term

|  | ESRGAN | SRIM | HyperRIM (ours) |
| --- | --- | --- | --- |
| IMLE-based |  | ✓ | ✓ |
| GAN-based | ✓ |  |  |
| Progressive Upscaling |  |  | ✓ |
| Hierarchical Sampling |  |  | ✓ |
| Intermediate Supervision |  |  | ✓ |
| Weight Normalization |  |  | ✓ |
| Has Mapping Network |  |  | ✓ |
| Has Pixel Loss | ✓ |  |  |
| Has Feature Loss | ✓ | ✓ | ✓ |

Table 4: Conceptual comparison among ESRGAN, SRIM and HyperRIM (ours).

## B. More Results

We include more $16\times$ super-resolution and image decompression results in the following pages. Videos of these results are also available as ancillary files on arXiv.

### B.1. $16\times$ Super-Resolution Results

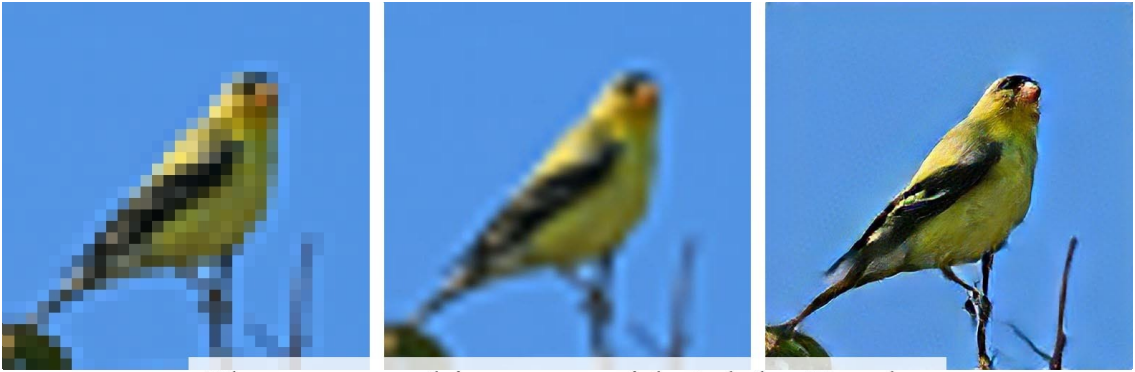
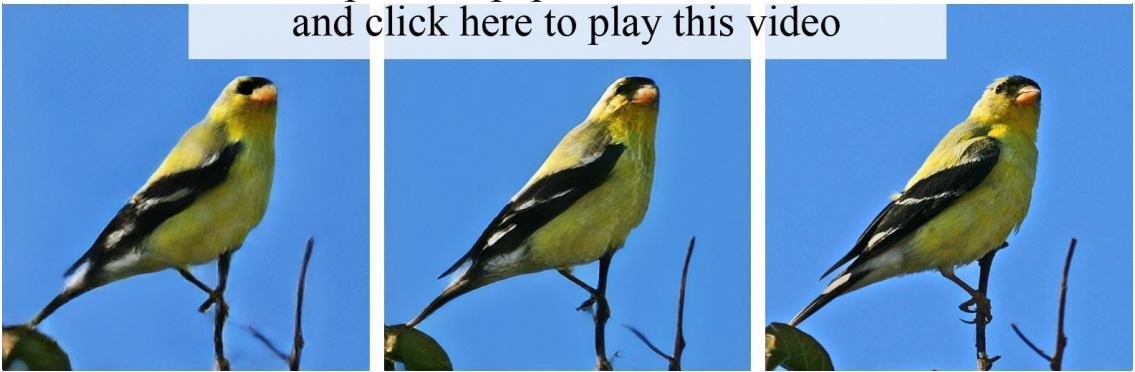

(a) Input     (b) Bicubic     (c) SRGAN

(d) SRIM     (e) HyperRIM     (f) Original Image

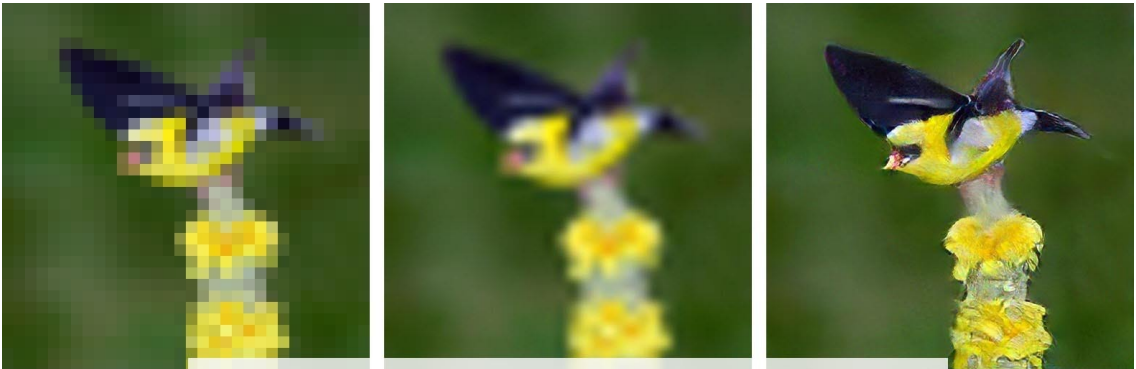
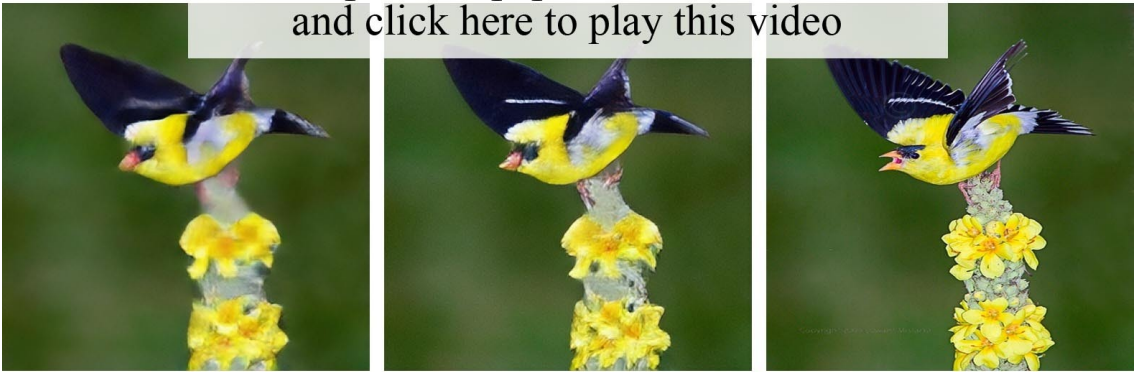

(a) Input     (b) Bicubic     (c) SRGAN

(d) SRIM     (e) HyperRIM     (f) Original Image

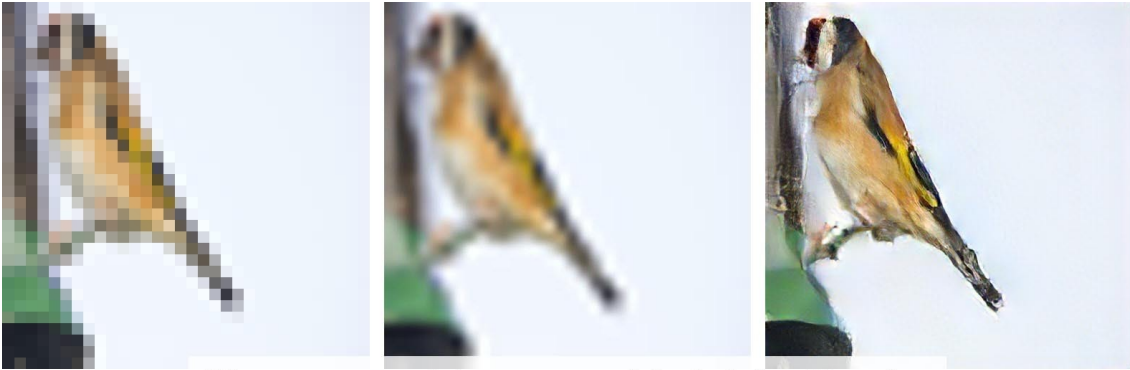
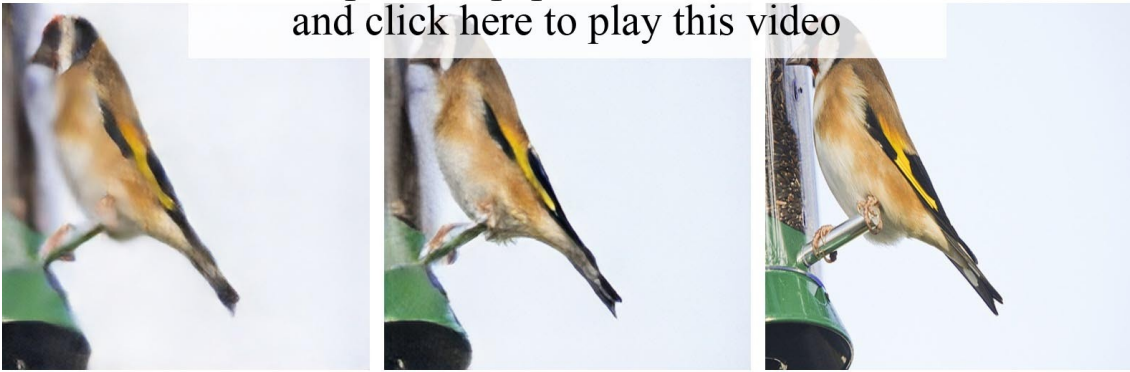

(a) Input (b) IE (c) SRGAN

(d) SRIM (e) HyperRIM (f) Original Image

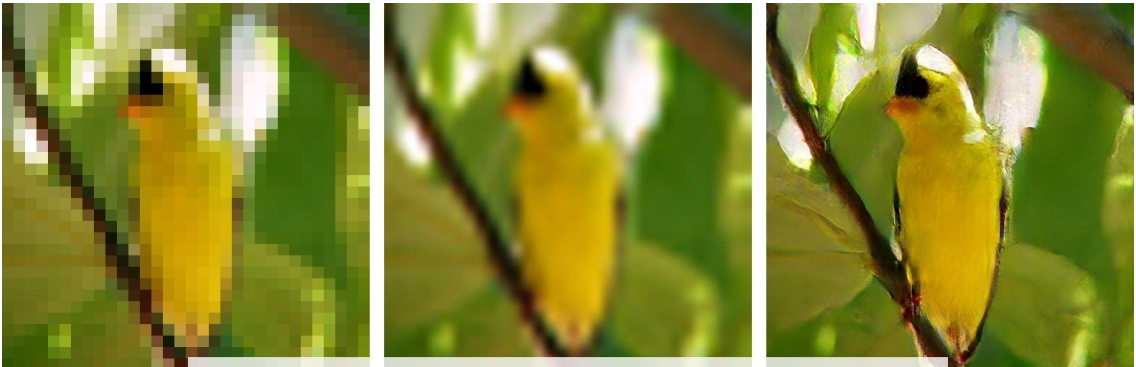
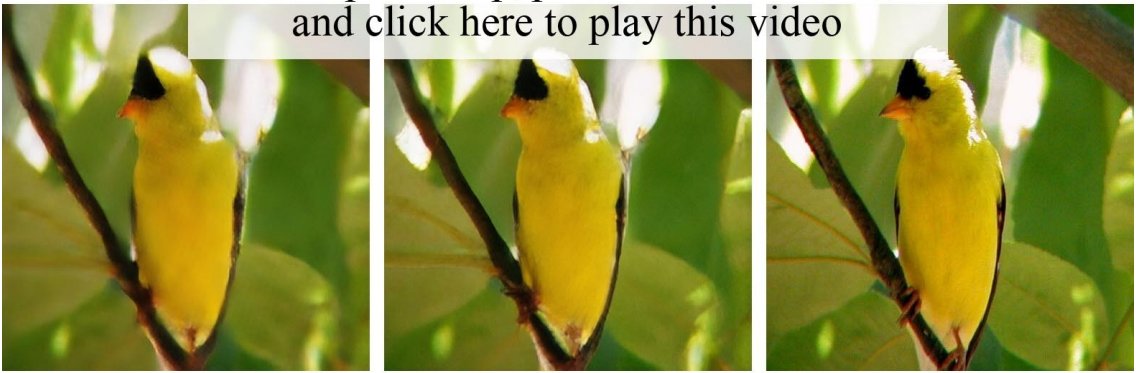

(a) Input (b) IE (c) SRGAN

(d) SRIM (e) HyperRIM (f) Original Image

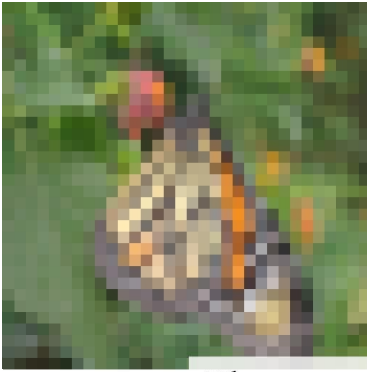 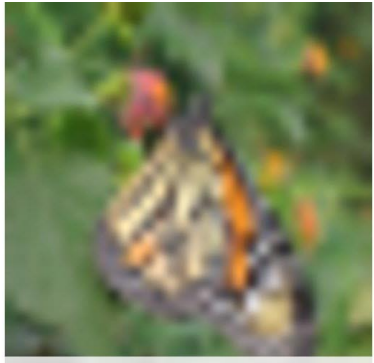 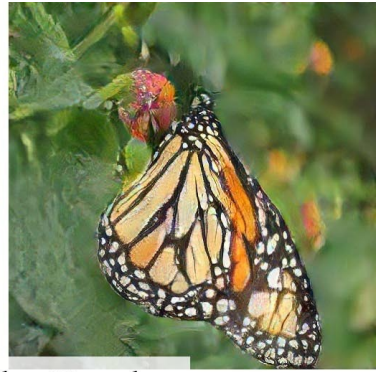

(a) Input      (b) Bicubic      (c) SRGAN

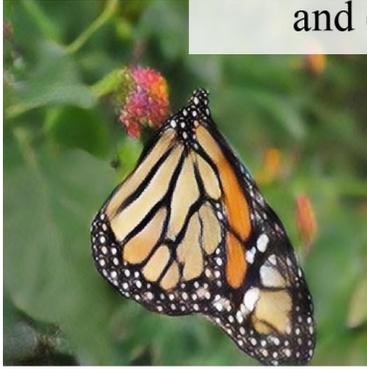 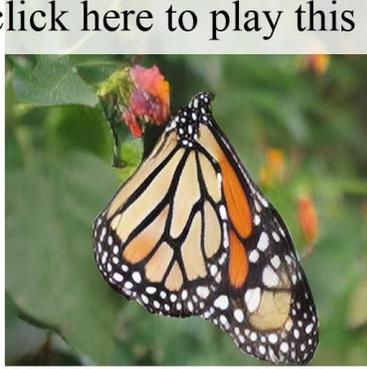 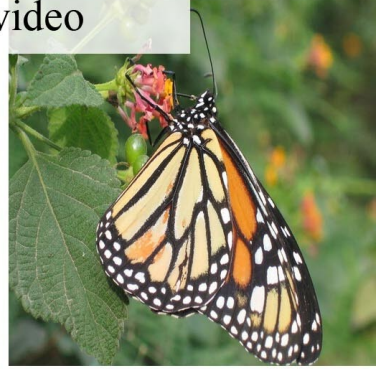

(d) SRIM      (e) HyperRIM      (f) Original Image

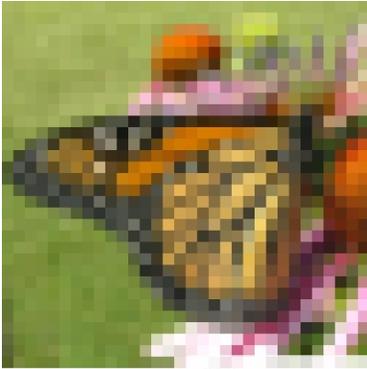 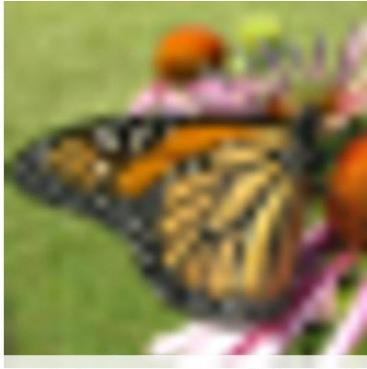 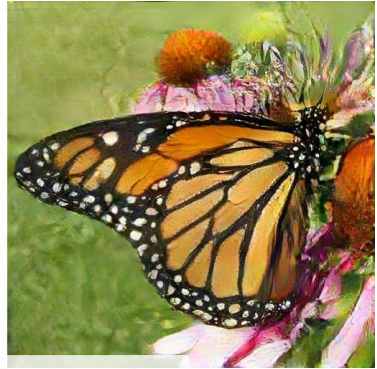

(a) Input      (b) Bicubic      (c) SRGAN

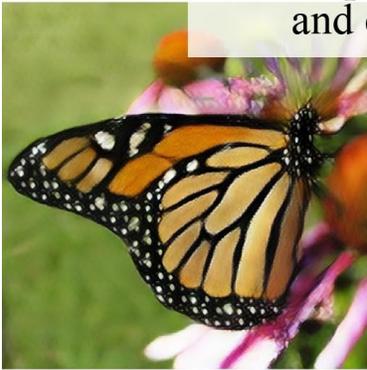 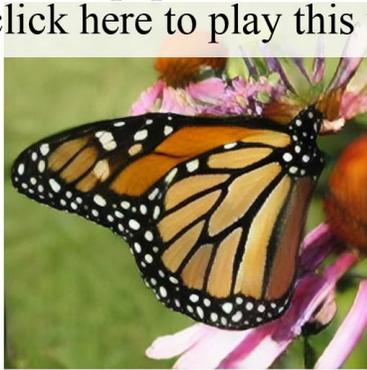 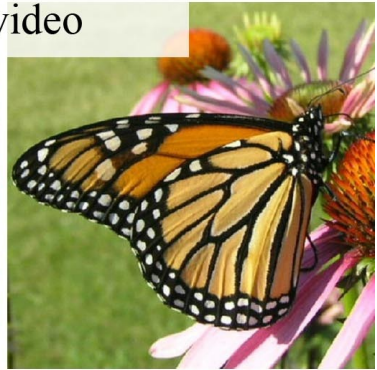

(d) SRIM      (e) HyperRIM      (f) Original Image

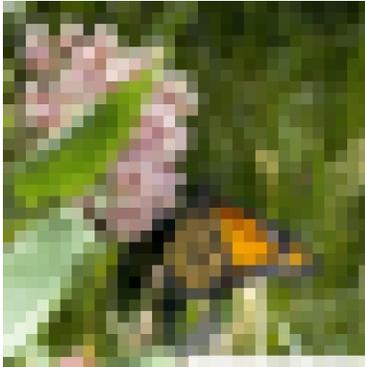 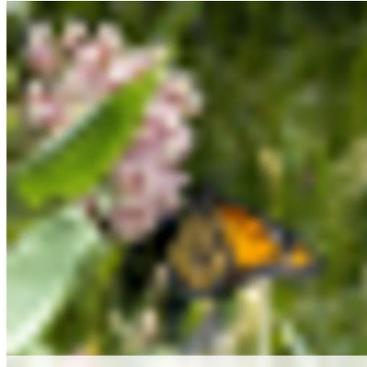 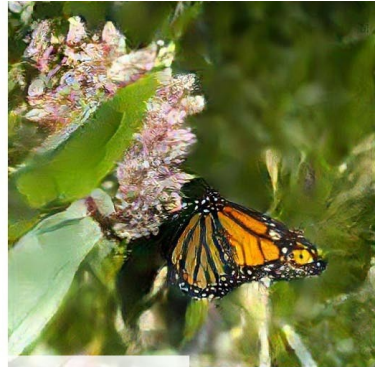

(a) Input        (b) Bicubic        (c) SRGAN

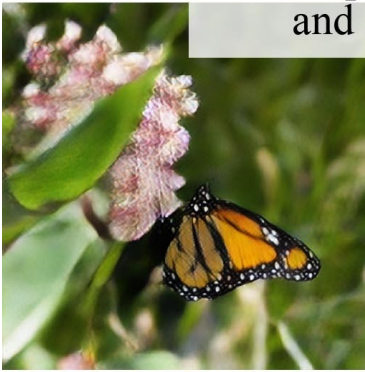 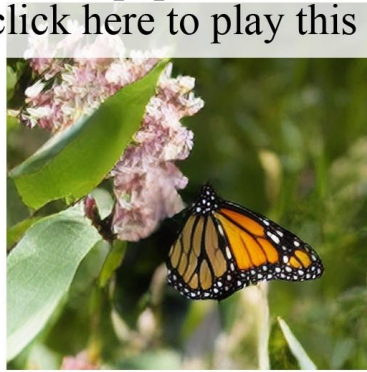 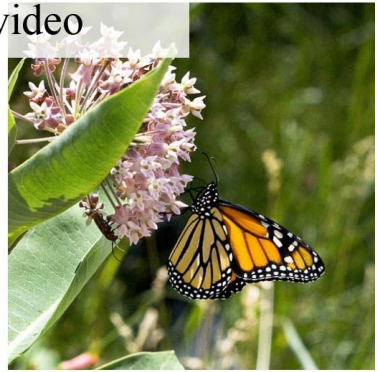

(d) SRIM        (e) HyperRIM        (f) Original Image

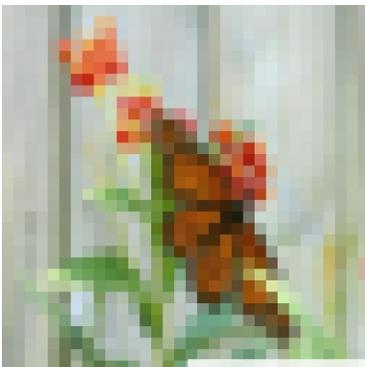 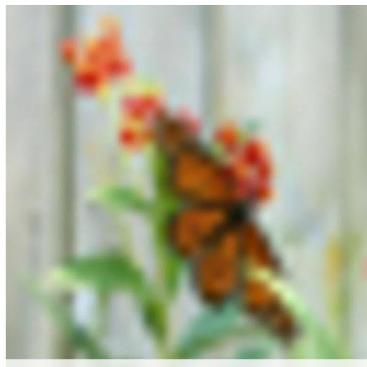 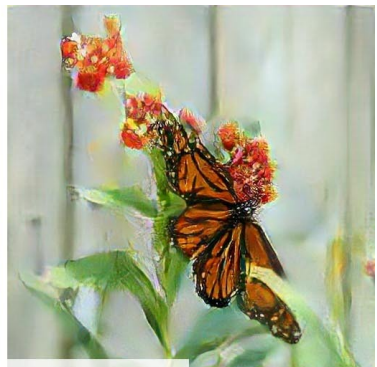

(a) Input        (b) Bicubic        (c) SRGAN

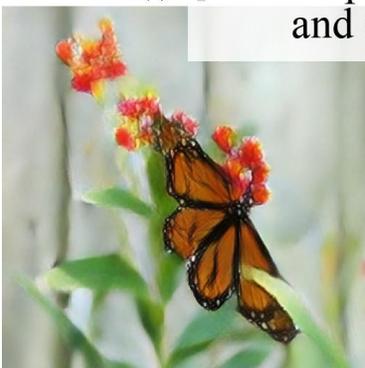 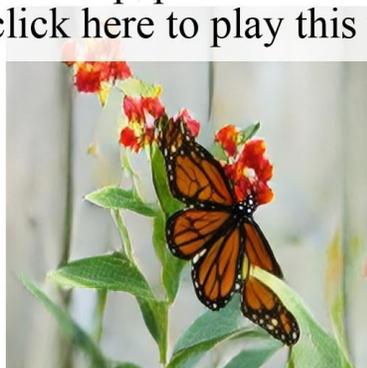 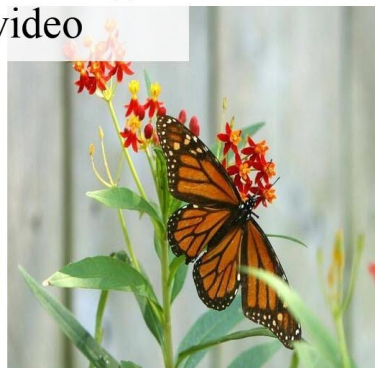

(d) SRIM        (e) HyperRIM        (f) Original Image

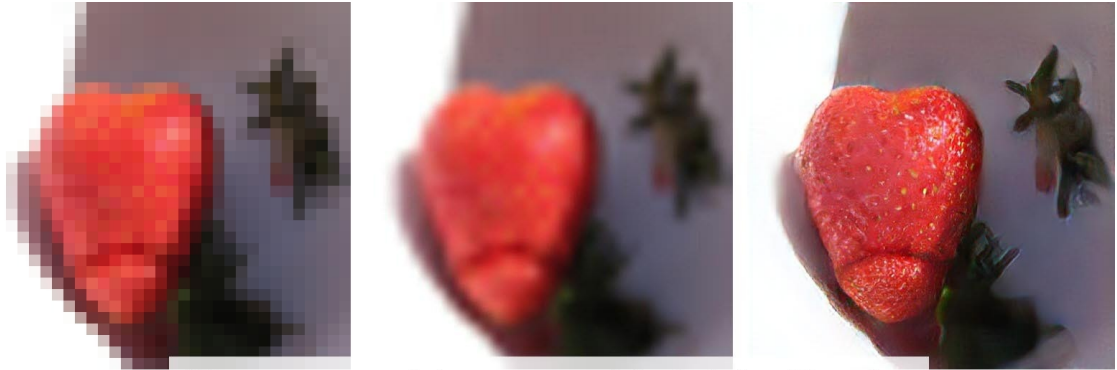
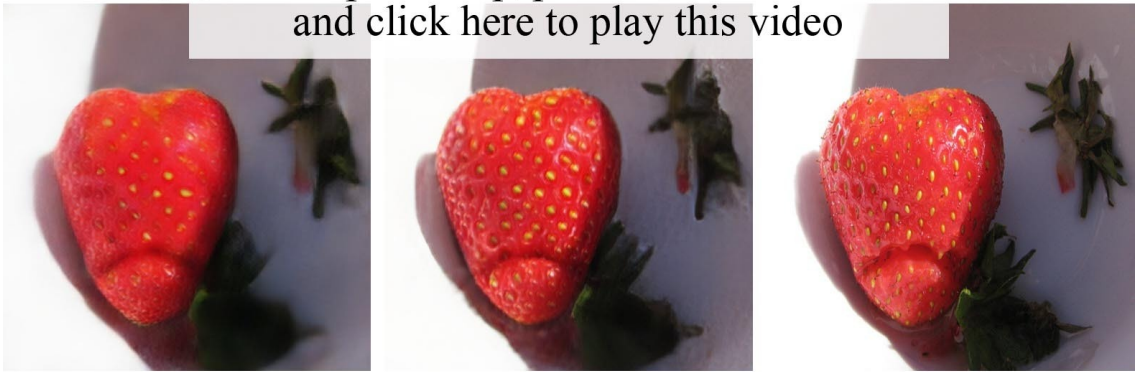

(a) Input     (b) Bicubic     (c) SRGAN

(d) SRIM     (e) HyperRIM     (f) Original Image

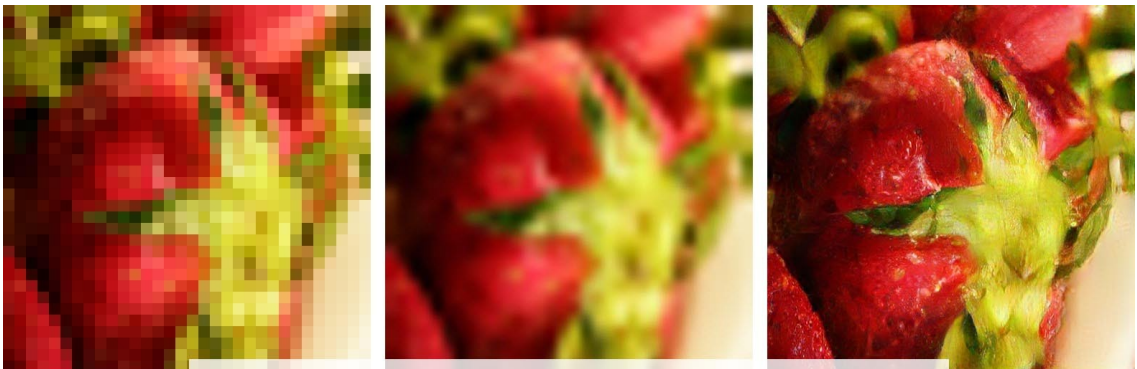
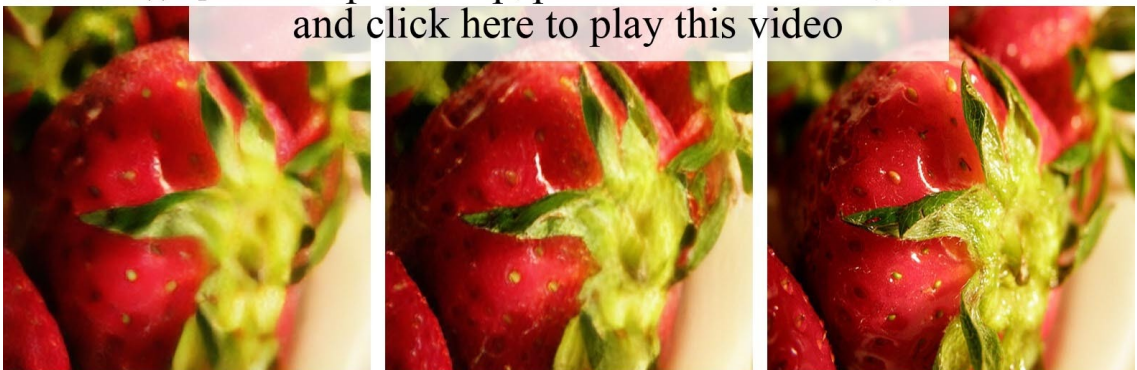

(a) Input     (b) Bicubic     (c) SRGAN

(d) SRIM     (e) HyperRIM     (f) Original Image

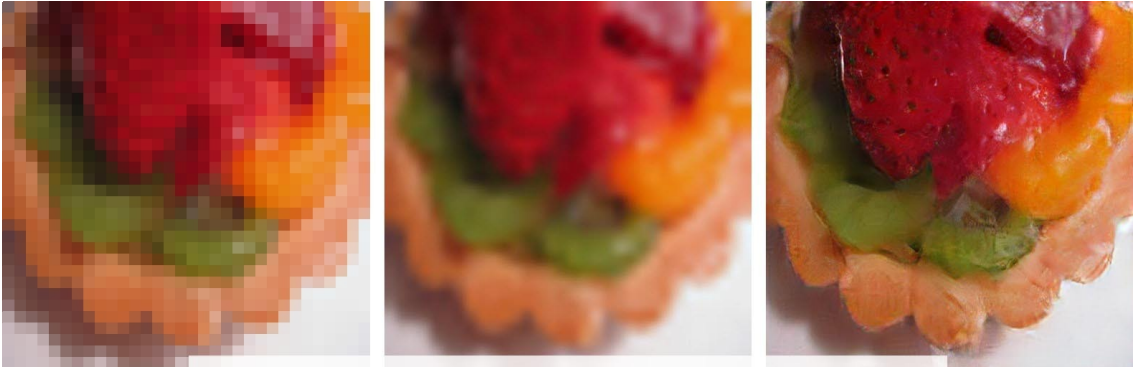
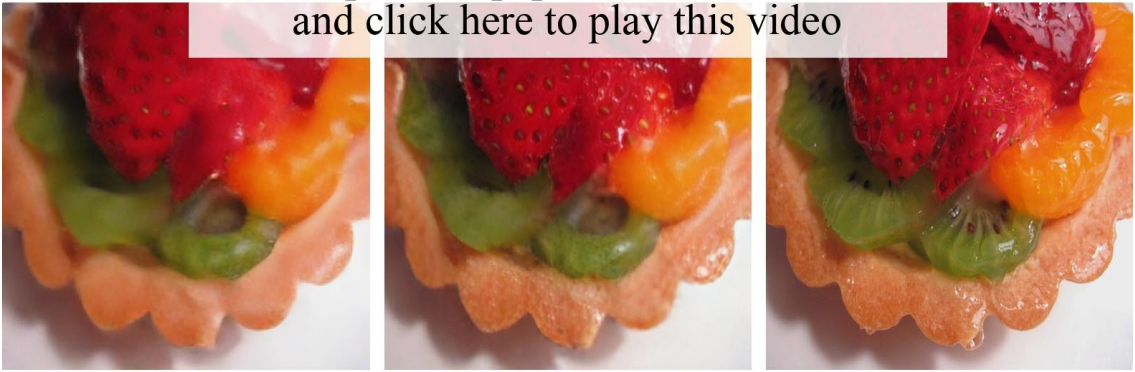

(a) Input     (b)     (c) SRGAN

(d) SRIM     (e) HyperRIM     (f) Original Image

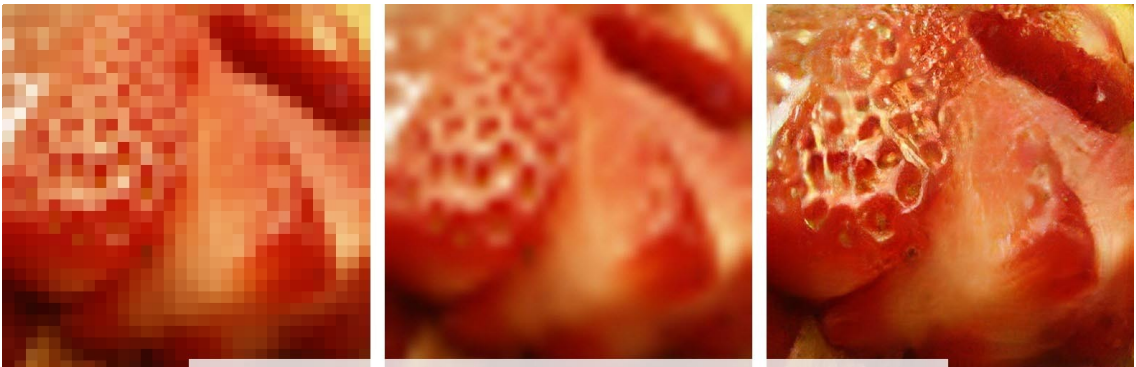
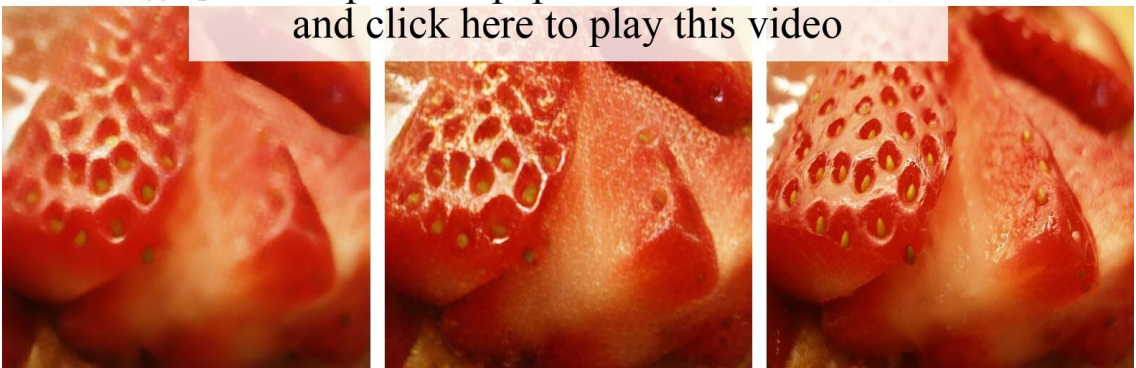

(a) Input     (b)     (c) SRGAN

(d) SRIM     (e) HyperRIM     (f) Original Image

## B.2. Image Decompression Results

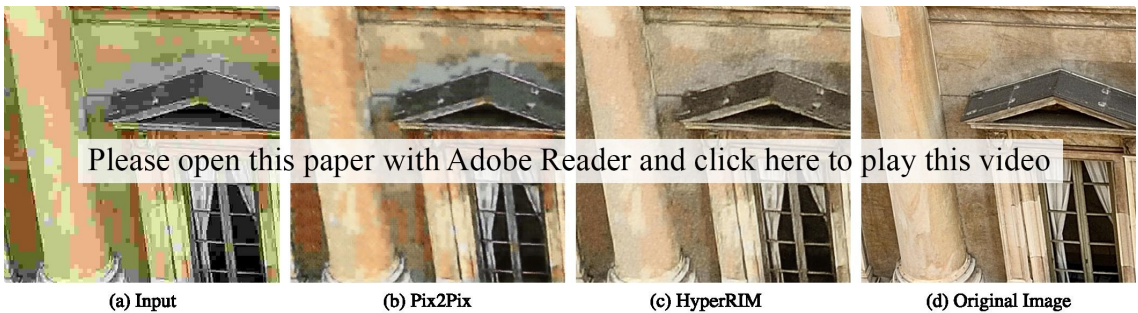
(a) Input (b) Pix2Pix (c) HyperRIM (d) Original Image

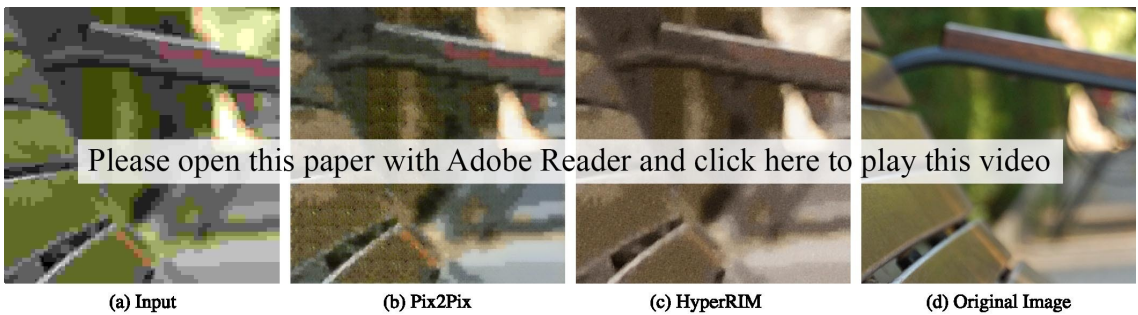
(a) Input (b) Pix2Pix (c) HyperRIM (d) Original Image

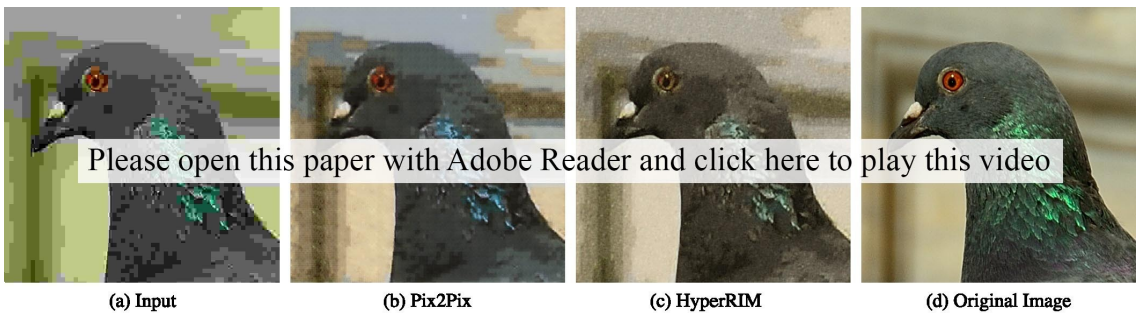
(a) Input (b) Pix2Pix (c) HyperRIM (d) Original Image

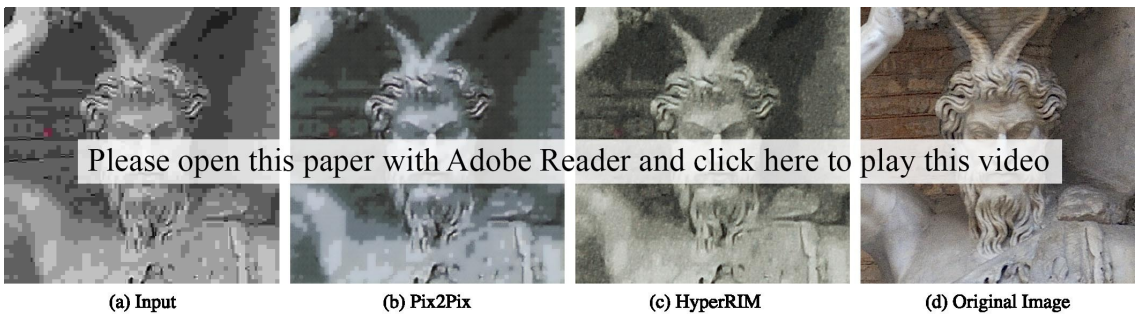
(a) Input (b) Pix2Pix (c) HyperRIM (d) Original Image